\documentclass{article}
\usepackage{arxiv}
\usepackage[numbers,sort&compress]{natbib}
\usepackage[utf8]{inputenc} 
\usepackage[T1]{fontenc}    
\usepackage{ dsfont }
\usepackage{hyperref}       
\usepackage{url}
\usepackage{textcomp}
\usepackage{ gensymb }
\usepackage[ruled,vlined]{algorithm2e}
\usepackage{algorithmic}
\usepackage{booktabs}       
\usepackage{amsfonts}       
\usepackage{nicefrac}       
\usepackage{microtype}      
\usepackage{lipsum}
\usepackage{float}
\usepackage{graphicx}
\usepackage{multirow}
\usepackage{array}
\usepackage{comment}
\usepackage{longtable}
\usepackage{ragged2e}
\usepackage{subfig}

\title{Defect Analysis of 3D Printed Cylinder Object Using Transfer Learning Approaches}

\author{
  Md Manjurul Ahsan \\
  School of Industrial and Systems Engineering\\
  University of Oklahoma\\
  Norman, Oklahoma-73019 \\
  \texttt{ahsan@ou.edu} \\
   \And
 Shivakumar Raman \\
  School of Industrial and Systems Engineering\\
  University of Oklahoma\\
  Norman, Oklahoma-73019\\
  \texttt{raman@ou.edu}\\
  \And
 Zahed Siddique \\
  School of Aerospace and Mechanical Engineering\\
  University of Oklahoma\\
  Norman, Oklahoma-73019\\
  \texttt{zsiddique@ou.edu}} 

\begin{document}
\maketitle
\begin{abstract}

Additive manufacturing (AM) is gaining attention across various industries like healthcare, aerospace, and automotive. However, identifying defects early in the AM process can reduce production costs and improve productivity - a key challenge. This study explored the effectiveness of machine learning (ML) approaches, specifically transfer learning (TL) models, for defect detection in 3D-printed cylinders. Images of cylinders were analyzed using models including VGG16, VGG19, ResNet50, ResNet101, InceptionResNetV2, and MobileNetV2. Performance was compared across two datasets using accuracy, precision, recall, and F1-score metrics. In the first study, VGG16, InceptionResNetV2, and MobileNetV2 achieved perfect scores. In contrast, ResNet50 had the lowest performance, with an average F1-score of 0.32. Similarly, in the second study, MobileNetV2 correctly classified all instances, while ResNet50 struggled with more false positives and fewer true positives, resulting in an F1-score of 0.75. Overall, the findings suggest certain TL models like MobileNetV2 can deliver high accuracy for AM defect classification, although performance varies across algorithms. The results provide insights into model optimization and integration needs for reliable automated defect analysis during 3D printing. By identifying the top-performing TL techniques, this study aims to enhance AM product quality through robust image-based monitoring and inspection.

\end{abstract}
\maketitle
\section{Introduction}

3D printing is a technology that has garnered considerable attention and popularity in recent years. It involves creating three-dimensional objects from a digital file by incrementally building up successive layers of material until the desired shape is achieved. This method enables the production of intricate and complex designs that would be challenging or impossible to fabricate using conventional manufacturing techniques~\cite{bhusnure20163d}. The range of materials suitable for 3D printing extends from plastics to metals, ceramics, and others. While initially employed primarily for prototyping and design purposes, 3D printing has since permeated various industries, including aerospace, automotive, medical, and fashion. The capacity to swiftly and effortlessly generate customized and specific objects have endowed 3D printing with significant utility in research and development, production processes, and even artistic and design endeavors~\cite{javan2018prototype}.

In recent years, the significance of 3D printing has been steadily growing, primarily owing to its capacity to swiftly generate highly customized and precise objects at a lower cost than traditional manufacturing methods~\cite{zou20233d}. This technological advancement introduces a novel dimension of creativity and innovation by enabling the rapid prototyping and production of intricate designs that were previously unattainable using conventional manufacturing techniques. Furthermore, 3D printing possesses the potential to bring about a revolutionary transformation in supply chain management and distribution by obviating the necessity for transporting large and cumbersome components, thereby promoting sustainability and environmental friendliness~\cite{khorasani2022review}. Additionally, it has found remarkable applications in developing prosthetics and medical implants, leading to personalized solutions in the healthcare domain~\cite{ramezani20234d}. Moreover, using 3D printing in space exploration has facilitated the on-demand production of parts and tools, significantly reducing the requirement for costly and time-consuming resupply missions~\cite{nazir2023multi}. As the evolution of 3D printing persists, its transformative impact is anticipated to extend across a broad spectrum of industries, reshaping conventional approaches to product creation and consumption~\cite{siddiqui2023emerging}.

Despite the many advantages of 3D printing, the technology still faces some challenges. One of the main challenges is the limited range of materials that can be used in 3D printing, particularly for industrial applications where specialized and high-performance materials are required~\cite{park20223d}. Another challenge is the limited print size of many 3D printers, making it difficult to produce larger objects~\cite{shahrubudin2020challenges}. The printing speed of 3D printers is also a challenge, as it can take hours or even days to print complex designs~\cite{dabbagh20223d}. Additionally, 3D printing requires a significant amount of energy and can generate a large amount of waste material~\cite{kantaros2021manufacturing}.

Moreover, the quality of the final product can be affected by the printing process, including the formation of defects such as warping and layer separation~\cite{lalegani2021overview}. Despite these challenges, researchers and engineers are working to overcome these limitations by developing new materials, improving printing techniques, and optimizing the printing process. The continued advancement of 3D printing technology is expected to address these challenges and unlock new possibilities for this revolutionary technology.  

Over the years, Machine Learning (ML) based approaches have shown promising results in addressing many of the challenges faced by 3D printing. One key application of ML in 3D printing is in predicting and preventing printing defects. By training ML models on large datasets of 3D printing data, it is possible to identify patterns and predict the likelihood of defects such as warping, shrinkage, and layer separation. This can help optimize the printing process and improve the quality of the final product~\cite{ansari2022layer}. Additionally, ML can optimize printing parameters, such as speed, temperature, and layer height, to reduce printing time and energy consumption while maintaining quality.
Furthermore, ML can be used to develop new materials optimized for 3D printing, allowing for a broader range of materials to be used in 3D printing~\cite{belei2022fused}. ML can also be used to improve the accuracy of 3D scanning and modeling, an essential step in designing and producing 3D printed objects. Overall, ML has the potential to significantly enhance the capabilities of 3D printing and overcome some of the challenges faced by this technology~\cite{haleem20213d}.

The current state of research on 3D printing defects is an active and growing area of interest for researchers and engineers. Defects in 3D printing can have significant implications for the quality and functionality of printed parts, making it essential to identify and prevent them. Researchers have investigated various defects, including warping, shrinkage, layer separation, and porosity. Many studies have focused on identifying the root causes of these defects, such as the printing parameters, material properties, and environmental conditions~\cite{mazzanti2019fdm}. Additionally, researchers have explored various methods for defect detection and prevention, including ML process optimization and real-time monitoring~\cite{mazzanti2019fdm}. However, despite the significant progress made in this field, many challenges remain to be addressed, such as the limited range of materials that can be used in 3D printing and the difficulty of detecting certain types of defects~\cite{paraskevoudis2020real}.

Several methods have been used in the past to identify 3D printing defects. One common approach is visual inspection, where printed parts are visually examined for defects such as warping, cracking, or uneven surfaces. However, this method is subjective and time-consuming, making it impractical for large-scale manufacturing~\cite{wang2022role}
Another approach is using microscopy, such as scanning electron microscopy (SEM) or optical microscopy, to examine the microstructure of printed parts and identify defects such as porosity or voids. This method provides high-resolution images that can reveal details not visible to the naked eye, but it is often destructive and can be limited by the size of the part being examined~\cite{demeneghi2021size}.

Other methods for defect identification include non-destructive testing techniques such as X-ray computed tomography (CT) or ultrasound, which can provide detailed internal images of printed parts to identify defects such as voids or delamination. However, these methods can be expensive and time-consuming, limiting their applicability for large-scale manufacturing~\cite{sun2022review}.

Transfer Learning (TL) is an ML technique that involves using a pre-trained model on one task to improve the performance of a related task. In TL, the knowledge gained from the pre-trained model is transferred to a new model being trained on a different but related task. This can significantly reduce the training data required to achieve high performance on the new task~\cite{ahsan2020study}.
In image classification, TL has been used to achieve state-of-the-art performance on various image recognition tasks. Convolutional neural networks (CNNs) are commonly used for image classification, and TL can be used to improve their performance. Some popular TL methods include pre-trained CNN, such as VGG, ResNet, or Inception, that has already been trained on a large dataset, such as ImageNet, to extract features from images. The pre-trained CNN is then fine-tuned on a new dataset for a different image recognition task, such as recognizing specific objects in medical images~\cite{ahsan2021detecting}. However, there are some limitations of the TL and CNN-based approaches in identifying the defect from the images of 3d printed cylinders, such as: 

\begin{itemize}
    \item The effectiveness of CNNs for 3D printing defect identification depends on the quality of the training data, and the dataset must be carefully curated to ensure that it is representative of the types of defects encountered in practice~\cite{wen2021application}.
    \item 3D printing defects can be challenging to detect, particularly internal defects such as voids or porosity that may not be visible on the surface of the printed part~\cite{dhakal2023impact}.
    \item Transfer learning requires a pre-trained model relevant to the specific application, and there may not be pre-trained models suitable for 3D printing defect detection~\cite{saeed2019automatic}.
    \item Interpreting the results of CNNs and transfer learning can be challenging, particularly for complex models, and it may be difficult to determine which features are most important for defect detection~\cite{liu2021defect}.
\end{itemize}
 	
This study aims to solve the classification of 3D-printed cylinder objects as either defective or non-defective using image classification techniques. The dataset used in this research is balanced, imbalanced, and collected from the Smart Materials and Intelligent Systems (SMIS) Laboratory, University of Oklahoma. This imbalance can pose a challenge to classification accuracy and can lead to biased results. Therefore, this research aims to accurately classify the objects and ensure that the classification is interpretable and explainable for researchers, practitioners, and users. This means that the model should be able to provide insights into how it arrived at its classification decisions and which features of the object are most important for classification. By addressing these challenges, this research can contribute to developing more reliable and interpretable image classification techniques for 3D printing defect detection, which can have important implications for improving the quality and efficiency of 3D printing processes.

The technical contribution of this study can be summarized as follows:

\section{Literature Review}
Over the years, several Deep Learning (DL) researchers have proposed based approaches to developing ML-based efficient 3d printing methods. For instance,  Saeed et al. (2019) propose an autonomous post-processor using CNN and DFF-NN algorithms to detect defects and estimate thermogram depth. They use pre-trained networks, fine-tuning, and pulsed thermography with CFRP samples. The approach enables real-time, automatic post-processing without human intervention, benefiting production lines. The study includes a comprehensive review of CNN's application to IR thermograms. Comparative results of different CNN models and object detection frameworks are presented. The proposed method demonstrates defect detection and depth estimation using 3D-printed CFRP samples. It could speed up inspections and integrate with quality control systems in production facilities~\cite{saeed2019automatic}.

Jia et al. (2022) propose a fabric defect detection system using transfer learning and an improved Faster R-CNN. The system addresses the limitations of existing algorithms in detecting small target defects with low accuracy. Transfer learning is employed by extracting pre-trained weights from the Imagenet dataset. ResNet50 and ROI Align replace VGG16 and ROI pooling layer for feature extraction. RPN and FPN are combined, using different IoU thresholds to distinguish positive and negative samples. The softmax classifier is utilized for image identification and predictions. The proposed system demonstrates superior accuracy and convergence ability compared to existing models~\cite{jia2022fabric}.

Chen et al. (2023) propose a deep learning approach for detecting low-contrast defects in 3D-printed ceramic curved surface parts. The method employs a blurry inpainting network model and a multi-scale detail contrast enhancement algorithm to improve image quality and enhance feature information. The detection of defects is performed using the ECANet-Mobilenet SSD network model. The proposed method achieves high accuracy in detecting crack and bulge defects, with an average detection time of 0.78s. However, the study focuses on only two types of defects and utilizes a specific dataset, which limits its generalizability. Overall, the study contributes to the advancement of intelligent defect detection in the advanced ceramic industry~\cite{chen2023defect}.

Li et al. (2022) present a comprehensive survey on Deep Transfer Learning (DTL) techniques in Intelligent Fault Diagnosis (IFD). The study covers the theoretical background of DTL and recent developments in IFD, offering recommendations for selecting DTL algorithms in practical applications and identifying future challenges. The survey aims to enhance researchers' understanding of state-of-the-art techniques and facilitate the design of effective solutions for IFD problems. However, empirical results or experiments are not provided, and the survey is limited to DTL-based IFD methods~\cite{li2023ifd}.

Ma et al. (2022) propose a transitive transfer learning CNN ensemble framework for bearing surface defect detection and classification. The framework addresses the limited dataset challenge in deep learning methods by incorporating a transfer path and transfer method selection strategy to enhance feature extraction. The proposed method achieves a high accuracy rate of 97.51\% and an average detection time of 155 ms per bearing, meeting industrial online detection requirements. However, the study lacks comparisons with other methods and focuses on only one type of defect. Overall, the framework shows promise for improving the industry's efficiency and accuracy of bearing quality inspection~\cite{ma2022novel}.

Lu et al. (2023) propose a system for real-time defect detection and closed-loop adjustment in additive manufacturing (AM) of carbon fiber reinforced polymer (CFRP) composites. The system utilizes a deep learning model for accurate real-time defect detection, classification, and evaluation. Defects are controlled through closed-loop adjustment of process parameters. The study's focus on CFRP composites limits direct applicability to other materials, and the system relies on accurate and accessible databases that may not always be available. Nevertheless, the proposed system demonstrates the potential of deep learning in real-time defect detection and closed-loop adjustment for AM processes. It accurately identifies two types of CFRP defects and quantifies their severity. The results highlight the effective control of defects through process parameter adjustments, offering a promising approach for enhancing the quality and efficiency of AM processes in CFRP composites~\cite{lu2023deep}.

Westphal (2022) introduces a novel machine-learning approach for quality assurance in Fused Deposition Modeling (FDM) additive manufacturing processes. The study utilizes environmental sensor data (temperature, humidity, air pressure, gas particles) and various state-of-the-art machine-learning algorithms to classify 3D printing conditions. A new data preparation method is developed, which sequences different sensor time series data. The XceptionTime architecture is identified as the most effective and robust against overfitting. A proof of concept comparing the ML analyses with 3D scan examinations in quality assurance reveals the ML analyses to be a more effective alternative in differentiating between various 3D printing conditions. The study is limited to FDM processes and only considers environmental sensor data for quality assurance. Overall, the findings present a novel and effective approach that has the potential to enhance the trust and acceptance of additive manufacturing processes in an industrial setting~\cite{westphal2022machine}.

Caggiano et al. (2019) propose a machine learning approach using a bi-stream deep convolutional neural network (DCNN) to identify defective conditions in Selective Laser Melting (SLM) of metal powders. The DCNN analyzes in-process images to recognize SLM defects, enabling adaptive process control and quality assurance. The bi-stream DCNN effectively combines layer images to classify defects caused by process non-conformities. Experimental evaluation demonstrates accurate defect recognition and classification, with potential applications in industrial quality assurance and adaptive process control for SLM. Further validation is needed for different systems and materials~\cite{caggiano2019machine}.

Wang et al. (2020) present a comprehensive review of machine learning (ML) applications in additive manufacturing (AM) across various domains. The review explores ML's potential in areas such as material design, topology design, process parameter optimization, in-process monitoring, and product quality assessment and control. Data security in AM is emphasized, and future research directions are outlined. While current ML applications in AM research primarily focus on processing-related processes, the review suggests a future shift towards new materials and rational manufacturing~\cite{wang2020machine}.

Qi et al. (2019) conducted a review on the utilization of neural networks (NNs) in additive manufacturing (AM) for complex pattern recognition and regression analysis. NNs offer a promising approach to address the challenge of constructing and solving physical models for the process-structure-property-performance relationship in AM. The review encompasses diverse application scenarios, including design, in-situ monitoring, and quality evaluation. Challenges related to data collection and quality control are identified, along with potential solutions. The authors emphasize the significant potential of combining AM and NNs to achieve agile manufacturing in the industry~\cite{qi2019applying}.

Kwon et al. (2020) applied a deep neural network to classify melt-pool images in selective laser melting based on 6 laser power labels. The neural network achieved satisfactory inference, even with images featuring blurred edges, by reducing the number of nodes with increasing layers. The proposed neural network demonstrated a low classification failure rate (under 1.1\%) for 13,200 test images and outperformed simple calculations in monitoring melt-pool images. The classification model could be utilized for non-destructive inference of unexpected alterations in microstructures or the identification of defective products. However, the study did not address potential limitations of using deep neural networks for classification in selective laser melting beyond its scope~\cite{kwon2020deep}.

Aquil et al. (2020) discuss software defect prediction and the application of machine learning techniques, including deep learning, ensembling, data mining, clustering, and classification, for predicting potential bugs during software maintenance. The study evaluates different predictor models using 13 software defect datasets and concludes that ensembling techniques consistently achieve high-accuracy predictions. The paper does not specify the types of software defects being predicted, and it is unclear if the datasets used for evaluation are representative of all software systems. The overall findings suggest that ensembling techniques are reliable for software defect prediction, but further research is needed to determine the optimal predictor model for different software systems~\cite{aquil2020predicting}.

Dogan et al. (2021) conduct a comprehensive literature review on the utilization of machine learning (ML) and data mining (DM) techniques in the field of manufacturing. The review categorizes existing solutions into scheduling, monitoring, quality, and failure topics. The paper highlights the benefits of employing ML techniques in manufacturing, presents statistical information on the field's current state, and suggests potential areas for future research. The knowledge discovery steps in the databases (KDD) process for manufacturing applications are explained in detail. However, the paper lacks a detailed analysis of the limitations and challenges associated with ML implementation in manufacturing~\cite{dogan2021systematic}.

Majeed et al. (2021) propose the BD-SSAM framework, which combines big data analytics, additive manufacturing, and sustainable smart manufacturing technologies to aid additive manufacturing enterprises in decision-making during the beginning of the product lifecycle. The framework is implemented in the fabrication of AlSi10Mg alloy components using selective laser melting (SLM) in additive manufacturing. The results indicate effective control over energy consumption and product quality, promoting smart, sustainable manufacturing, emission reduction, and cleaner production. However, the study's focus is limited to the beginning of the product lifecycle and specific materials (AlSi10Mg alloy components with SLM). Further research is needed to validate the framework across different materials and stages of the product lifecycle. Overall, the paper presents a promising framework that integrates big data analytics, additive manufacturing, and sustainable smart manufacturing to support smart, sustainable manufacturing and cleaner production, particularly during the beginning of the product lifecycle in the additive manufacturing industry~\cite{majeed2021big}.

Azamfar et al. (2020) introduce a deep learning-based domain adaptation method for fault diagnosis in semiconductor manufacturing. The method aims to mitigate variations in manufacturing processes that can impact data-driven approaches. It utilizes a deep neural network and the maximum mean discrepancy metric to optimize high-level data representation. Experimental results on a real-world semiconductor manufacturing dataset demonstrate the effectiveness and generalization of the proposed method for quality inspection. However, the paper does not discuss the limitations of the proposed method~\cite{azamfar2020deep}.

Qu et al. (2019) present an overview of smart manufacturing systems (SMSs), which have gained significant attention from countries and manufacturing enterprises. The paper covers the evolution, definition, objectives, functional requirements, business requirements, technical requirements, and components of SMSs. An autonomous SMSs model is also proposed driven by dynamic demand and key performance indicators~\cite{qu2019smart}. However, the paper lacks a discussion on the challenges and limitations of implementing SMSs and potential risks and concerns related to data privacy and security. The findings provide a comprehensive understanding of SMSs and offer a reference for the transformation of manufacturing enterprises to intelligent systems.

Based on the overall analysis, it can be observed that the majority of studies have focused on employing machine learning (ML) based approaches to classify or identify defects and faults in both the 3D printed objects and the 3D printing process. While most of the studies have demonstrated the potential of convolutional neural network (CNN) based approaches, only a limited number of studies have explored transfer learning (TL) based approaches, which offer computational efficiency and applicability in identifying defects across various sizes and types of 3D printed objects. Furthermore, conducting an analysis of TL-based approaches for defect analysis using images obtained from 3D-printed objects will enable us to determine the feasibility of using TL-based approaches as viable solutions for developing ML-based defect analysis on a large scale.

However, none of the existing studies offer sufficient contextual information to assist future researchers and practitioners in comprehending whether there is an ascertainable impact on their methodologies that can be elucidated. Consequently, there arises a necessity for a comprehensive study capable of delivering novel insights and facilitating an enhanced understanding of the characteristics inherent in machine learning (ML)-based models and their interpretability concerning the identification of defect analysis in additive manufacturing.

\section{Motivation}
The motivation behind this study is to evaluate the performance of TL approaches in the context of defect analysis in Additive Manufacturing (AM). With the advancements in AM technology, detecting defects during the printing process has become an essential issue to ensure the quality of the printed product. Image classification using TL models has shown promising results in various image-based applications. However, in the context of AM, challenges include imbalanced data, variability in printing conditions, and complexity in the defect patterns~\cite{westphal2022machine, zeiser2023data}. Therefore, evaluating the performance of TL models, such as VGG16, VGG19, ResNet50, ResNet101, and MobileNetV2, is essential to determine their effectiveness in detecting defects in 3D printed products. 

Therefore, this study aims to evaluate and compare the performance of various TL models on a dataset of 3D-printed product images to determine the best approach for defect analysis in additive manufacturing. The study aims to provide insights into the effectiveness of TL models and their potential contribution to the field of additive manufacturing. By identifying the best TL model for defect analysis, this research is expected to enhance the quality of 3D-printed products and promote the advancement of the field.

The remainder of the paper is organized as follows: The methodology of the experiment is described in detail in Section~\ref{methods}. The results of the experiment are presented in Section~\ref{observation}. A brief discussion of the study's findings is provided in Section~\ref{discussion}, and the overall conclusions of the research and potential avenues for future research are summarized in Section~\ref{conclusions}.
\section{Methodology}\label{methods}
This section outlines the procedure for acquiring the 3D printing images, the TL model used in this study, the experimental setup, and the evaluation metrics employed to assess the model's performance.
\subsection{Data Collection}
\begin{figure}[H]
    \includegraphics[width=\textwidth]{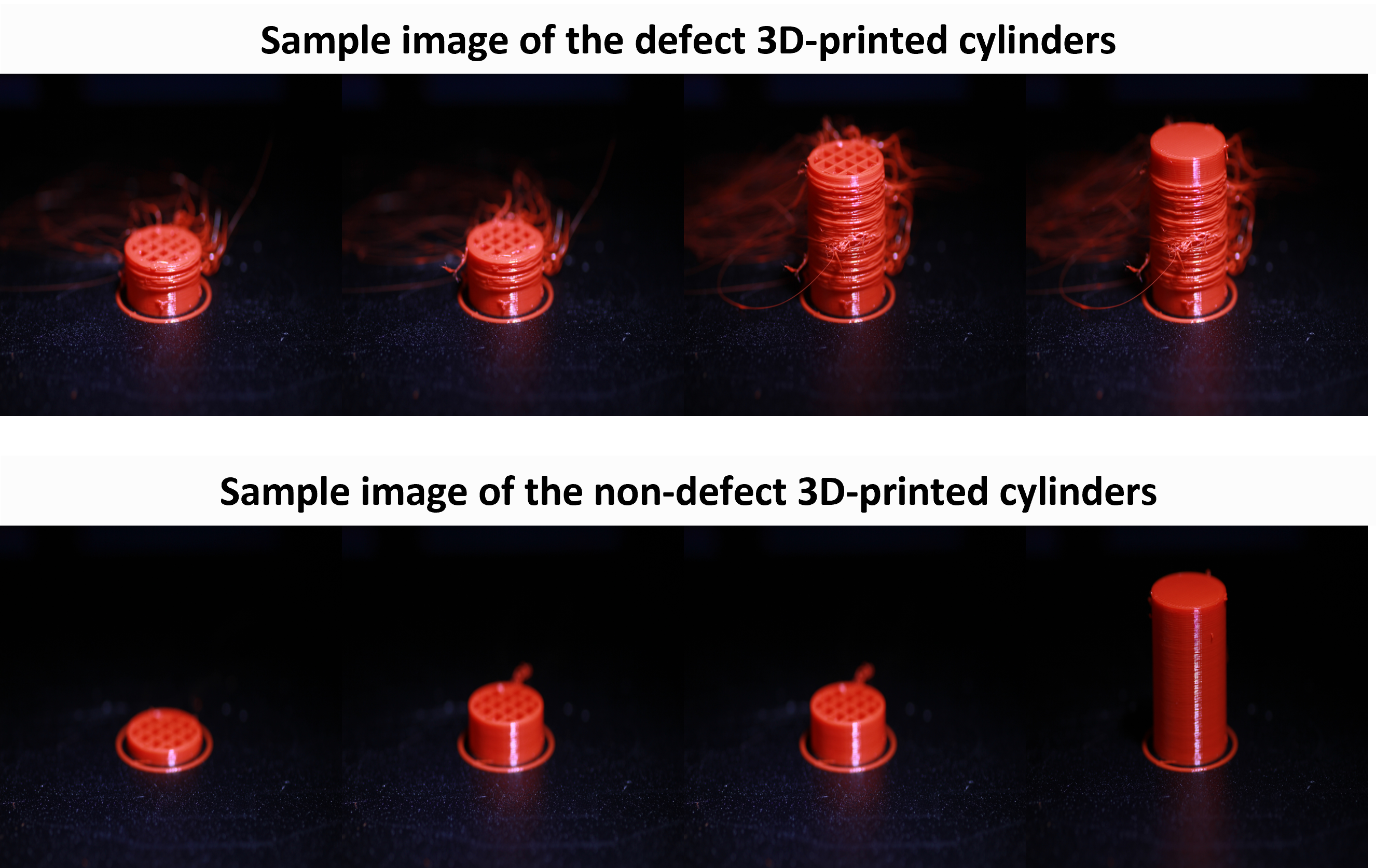}
    \caption{Representative samples of 3D-printed product images obtained from SMIS, which have been utilized in this research.}
    \vspace{-\baselineskip}
    \label{fig:3ds}
\end{figure}

The dataset was generously provided by the Smart Materials and Intelligent Systems (SMIS) Laboratory. It comprises a collection of 3D printed prototype images of cylindrical objects, some of which contain defects while others are non-defective.
Figure~\ref{fig:3ds} displays a selection of typical images of cylindrical objects from the dataset, including both defective and non-defective examples. Table~\ref{tab:3d} provides an overview of the allocation of data for training and testing of each of the CNN models that were investigated. In both studies, six different DL approaches were investigated: VGG16~\cite{simonyan2014very}, InceptionResNetV2~\cite{szegedy2017inception}, ResNet50~\cite{akiba2017extremely}, MobileNetV2~\cite{sandler2018mobilenetv2}, ResNet101~\cite{he2016deep} and VGG19~\cite{simonyan2014very}.

\begin{table}[H]
\centering
\caption{Allocation of data for training and testing of TL models used in this study.}
\label{tab:3d}
\resizebox{.5\textwidth}{!}{%
\begin{tabular}{@{}lllll@{}}
\toprule
Dataset & Label & Train & Test & Total \\ \midrule
\multirow{2}{*}{Study One} & Defect & 105 & 31 & 136 \\
 & Non-defect & 112 & 24 & 136 \\
\multirow{2}{*}{Study Two} & Defect & 736 & 211 & 947 \\
 & Non-defect & 1677 & 393 & 2070  \\ \bottomrule
\end{tabular}%
}
\end{table}

\subsection{Deep Learning Algorithms}
\subsubsection{VGG}
The Visual Geometry Group (VGG), proposed by the Visual Geometry Group, introduced two CNN models known as VGG16 and VGG19. These models are characterized by their exceptionally deep architecture, consisting of 16 and 19 layers, respectively. To train these models, a dataset comprising one million samples from the ImageNet dataset was utilized. Specifically, the VGG16 model is designed to process input images with dimensions of 224 by 224 pixels.

During the forward pass, images are subjected to a series of Convolutional layers featuring filters ranging from 64 to 512. The pooling layer is implemented using a 2-by-2 filter with max-pooling. The Rectified Linear Unit (ReLU) serves as the activation function with a stride size of 2. Subsequently, the pooling layer output is transformed into a 1D vector within the dense layer. Ultimately, the Softmax activation function is applied to the output layer, facilitating the classification of samples into one of the 1000 available categories~\cite{simonyan2014very}.
\subsubsection{InceptionResNetV2}
In 2017, a deep learning architecture known as InceptionResNetv2 was proposed by Szegedy et al. This architecture combines the strengths of both the Inception and ResNet architectures, resulting in synergistic advantages for image classification tasks. With a total of 164 layers, IRv2 employs 1x1, 3x3, and 5x5 convolutions, in addition to max pooling, to extract features from input images. By incorporating residual connections, the architecture enables the training of deeper networks while effectively addressing the issue of vanishing gradients. Furthermore, batch normalization and the Rectified Linear Unit (ReLU) activation function are integrated into the architecture, thereby enhancing the speed and accuracy of the training process.

Significantly, IRv2 achieved remarkable performance on the ImageNet Large Scale Visual Recognition Challenge dataset in 2016, attaining a top-5 accuracy of 95.3\%~\cite{szegedy2017inception}.
\subsubsection{ResNet}
ResNet, short for Residual Network, is a deep neural network architecture introduced by researchers at Microsoft in 2015 to tackle the challenge of vanishing gradients in extremely deep neural networks. ResNet50, a variant of ResNet, implements the model with 48 convolution layers, incorporating one layer each of max pooling, average pooling, and regular pooling. The architecture of ResNet50 begins with a 7x7 convolutional layer consisting of 64 kernels, each with a stride size of 2. This is followed by a max-pooling layer with a stride size of 2. The network then employs nine convolutional layers with kernel filters of sizes 64, 64, and 256, respectively. The subsequent nine levels of the network comprise kernel filters of sizes 512, 512, and 2048.

An FC (fully connected) layer with 1000 nodes, along with an average pooling layer, is added, and the output layer employs the Softmax function as its activation function. ResNet101 shares a similar architecture to ResNet50, but it includes an additional three-block layer within the fourth block, consisting of 256, 256, and 1024 filters~\cite{he2016deep,ahsan2023deep,ahsan2022transfer}.
\subsubsection{MobileNetV2}
MobileNetV2 is an advanced module that incorporates inverted residuals and a linear bottleneck. This efficient model is specifically designed for utilization in mobile phone applications, where it can accommodate low-dimensional inputs, minimize computational operations, and consume less memory while maintaining a high level of accuracy. MobileNetV2 leverages depthwise separable convolution, which divides the convolution operation into two distinctive layers.

The initial layer is the depth-wise convolution, which excels in efficiency by performing a single filtering operation. The subsequent layer captures additional features through the linear computation of the input. By employing MobileNetV2, the computational cost of standard model layers has been significantly reduced by a factor of $k^2$. In many cases, MobileNetV2 saves approximately 8 to 9 times the computational resources required for a 3x3 depth-wise separable convolution compared to traditional models~\cite{sandler2018mobilenetv2}.

\subsection{Using Pre-Trained Convet}

A pre-trained network refers to a network that has been previously trained on a larger dataset, enabling it to acquire a comprehensive hierarchy for feature extraction. This attribute makes pre-trained networks particularly effective when dealing with smaller datasets. An exemplary instance of such a network is the VGG16 architecture, which was developed by Simonyan and Zisserman in 2014~\cite{simonyan2014very}. Figure 1 showcases a sample architecture illustrating the procedure of the pre-trained model. It is worth noting that all the models employed in this study are readily available as pre-packaged options within the Keras framework~\cite{chollet2017deep}.

The fine-tuning process on the VGG16 network is carried out according to the following steps:

\begin{enumerate}
\item Initially, the models are initialized with a pre-trained network that lacks a Fully Connected (FC) layer.
\item Subsequently, a new connected layer is introduced, incorporating a pooling layer and applying the "softmax" activation function. This newly created layer is appended on top of the VGG16 model.
\item Finally, during the training phase, the weights of the convolutional layers are frozen to ensure that only the FC layer undergoes training for the duration of the experiment.
\end{enumerate}

The identical procedure was employed for all other deep learning techniques. For each CNN architecture utilized in the experiment, the model was further modified according to the following construction:

$AveragePooling2D(Pool size=(4,4))\rightarrow Flatten\rightarrow Dense \rightarrow  Dropout (0.5) \rightarrow Dense (Activation=``softmax")$.
\begin{figure}[H]
    \centering
    \includegraphics[width=\textwidth]{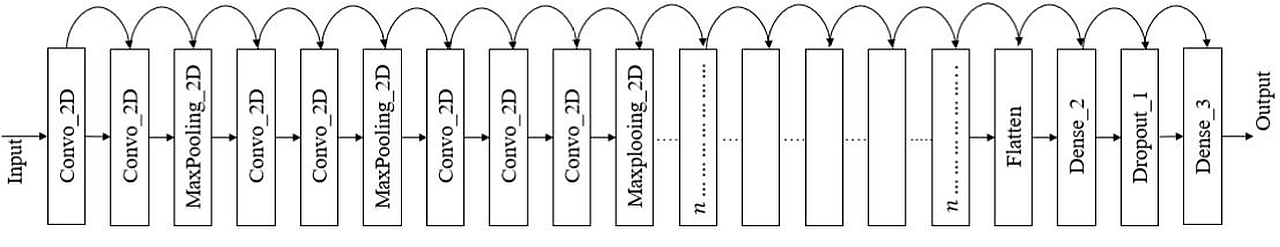}
    \caption{VGG16 architecture used during this experiment.}
    \label{fig:vgg16}
\end{figure}

As is well known, most pre-trained models consist of multiple layers associated with various parameters, such as the number of filters, kernel size, hidden layers, and neurons~\cite{denil2013predicting}. However, manually fine-tuning these parameters can be a time-consuming process~\cite{hutter2015beyond, qolomany2017parameters}. Taking this into consideration, three parameters of the deep learning model are optimized in this study: batch size (which determines the number of samples processed before updating internal model parameters), epochs (indicating the number of times the learning algorithm iterates over the entire dataset), and learning rate (a hyperparameter controlling the magnitude of model adjustments when updating model weights to calculate error)~\cite{brownlee2018difference}.

Inspired by prior works~\cite{smith2018disciplined,smith2017don}, a grid search method commonly employed for parameter tuning was applied in this study~\cite{bergstra2012random}. Initially, the following parameters were randomly selected:
\begin{center}
Batch\, size = $[4,5,8,10]$\\
Number\, of\, epochs = $[10,20,30,40]$\\
Learning\, rate = $[.001, .01, 0.1]$
\end{center}

For Study One, better results were achieved using the grid search method with the following parameters:

\begin{center}
Batch\, size = $8$\\
Number\, of\, epochs = $30$\\
Learning\, rate = $.001$\\
\end{center}
Similarly, for Study Two, the best results were achieved with:
\begin{center}
Batch\, size = $50$\\
Number\, of\, epochs = $50$\\
Learning\, rate = $.001$\\
\end{center}

The optimization algorithm employed for all models was Adam, an adaptive learning rate optimization algorithm recognized for its robust performance in binary image classification\cite{perez2017effectiveness,filipczuk2013computer}. Adhering to the customary procedure in data mining techniques, 80\% of the accessible data was designated for training, while the remaining 20\% was set aside for testing purposes~\cite{mohanty2016using, menzies2006data, stolfo2000cost}. Each study was replicated twice, and the ultimate outcome was established by calculating the average of the outcomes from the two independent experiments, aligning with the guidance put forth by Zhang et al. (2020)~\cite{zheng2020viral}. 
\subsection{Experimental Setup}

The entire experiment was conducted using an office-grade laptop equipped with standard specifications, which encompassed a Windows 10 operating system, an Intel Core i7-7500U processor, and 16 GB of RAM. The assessment of performance was conducted utilizing diverse metrics, encompassing model accuracy, precision, recall, and F1-score~\cite{ahsan2020face}. 

\begin{equation}\label{1}
   \textrm{Accuracy} =\frac{t_p + t_n}{t_p + t_n + f_p + f_n} 
\end{equation}

\begin{equation}\label{2}
 \textrm{Precision} = \frac{t_p}{t_p+f_p}   
\end{equation}

\begin{equation}\label{3}
  \textrm{Recall}=\frac{t_p}{t_n+f_p}  
\end{equation}

\begin{equation}\label{4}
    \textrm{F1-score} =2\times\frac{\textrm{Precision}\times\textrm{ Recall}}{\textrm{Precision+Recall}}
\end{equation}

Where,
\begin{itemize}
\item True Positive ($t_p$)= Defect cylinder classified as Defect
\item False Positive ($f_p$)= Normal cylinder classified as Defect
\item True Negative ($t_n$)= Normal cylinder classified as Normal cylinder
\item False Negative ($f_n$)= Defect cylinder classified as Normal cylinder.
\end{itemize}

\section{Results}\label{observation}

\subsection{Study One}
Table~\ref{tab:std1} presents the performance of six different DL algorithms, namely VGG16, InceptionResNetV2, ResNet50, ResNet101, MobileNetV2, and VGG19, on the train set. The highest performance in terms of all four metrics was achieved by VGG16 and MobileNetV2, both with perfect scores. InceptionResNetV2 also performed well, averaging 99\% across all metrics. On the other hand, ResNet50 had the lowest performance among the models, with an average F1-score of 0.34. 
\begin{table}[H]
\centering
\caption{Computational results of the various Transfer Learning models used in Study One on the train set.}
\label{tab:std1}
\resizebox{\textwidth}{!}{%
\begin{tabular}{@{}lcccc@{}}
\toprule
Algorithm & \multicolumn{1}{l}{Accuracy} & \multicolumn{1}{l}{Precision} & \multicolumn{1}{l}{Recall} & \multicolumn{1}{l}{F1-score} \\ \midrule
VGG16 & 1 & 1 & 1 & 1 \\
InceptionResNetV2 & .99 & .99 & .99 & .99 \\
ResNet50 & 0.51 & 0.25 & 0.5 & 0.34 \\
ResNet101 & 0.86 & 0.87 & 0.85 & 0.85 \\
MobileNetV2 & 1 & 1 & 1 & 1 \\
VGG19 & 0.98 & 0.98 & 0.98 & 0.98 \\ \bottomrule
\end{tabular}%
}
\end{table}

\begin{figure}[H]
    \centering
    \includegraphics[width=\textwidth]{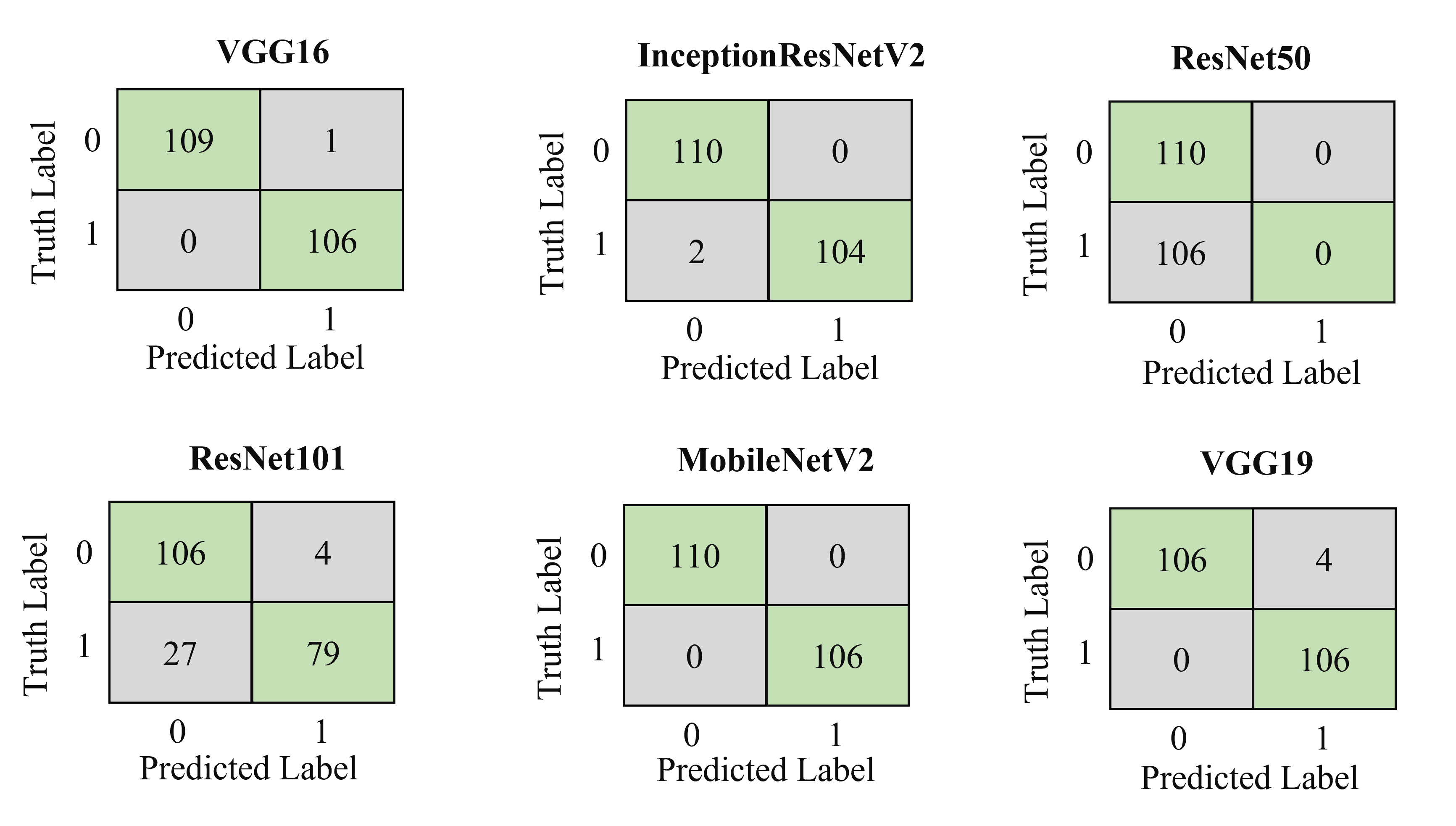}
    \caption{Confusion matrices of the different Transfer Learning models used during Study One on the train set.}
    \label{fig:cm1a}
\end{figure}

The confusion matrix on the train data, as shown in Figure~\ref{fig:cm1a} for Study One, indicates the true positive, false positive, false negative, and true negative values for each algorithm. The results show that VGG16, InceptionResNetV2, and MobileNetV2 had perfect classification accuracy, with 109, 110, and 110 true positive values, respectively. ResNet50 had the highest number of false negative values, with 106, and a relatively low number of true positives, with only 110. ResNet101 had 27 false negative values but a higher number of true positives than ResNet50, with 106. Finally, VGG19 had the same number of false negatives as ResNet50, with 0, but a higher number of false positives, with 4. Overall, the confusion matrix on the train data shows that VGG16, InceptionResNetV2, and MobileNetV2 performed the best, while ResNet50 and VGG19 had lower accuracy due to higher false negative and false positive values, respectively.

Table~\ref{tab:std1test} presents the results of Study One model performance on the test set. The table shows that VGG16, InceptionResNetV2, and MobileNetV2 achieved perfect scores of 1 for all evaluation metrics, while ResNet50 had the lowest accuracy, precision, and F1-score. ResNet101 achieved high performance with an accuracy of 0.85 and an F1-score of 0.85, while VGG19 had a high accuracy of 0.98 and an F1-score of 0.98. These results demonstrate that different TL algorithms perform differently on the same dataset.
\begin{table}[H]
\centering
\caption{Computational results of the various Transfer Learning models used in Study One on the test set.}
\label{tab:std1test}
\resizebox{\textwidth}{!}{%
\begin{tabular}{@{}lllll@{}}
\toprule
Algorithm & Accuracy & Precision & Recall & F1-score \\ \midrule
VGG16 & 1 & 1 & 1 & 1 \\
InceptionResNetV2 & 1 & 1 & 1 & 1 \\
ResNet50 & 0.47 & 0.24 & 0.50 & 0.32 \\
ResNet101 & 0.85 & 0.87 & 0.86 & 0.85 \\
MobileNetV2 & 1 & 1 & 1 & 1 \\
VGG19 & 0.98 & 0.98 & 0.98 & 0.98 \\ \bottomrule
\end{tabular}%
}
\end{table}

\begin{figure}
    \centering
    \includegraphics[width=\textwidth]{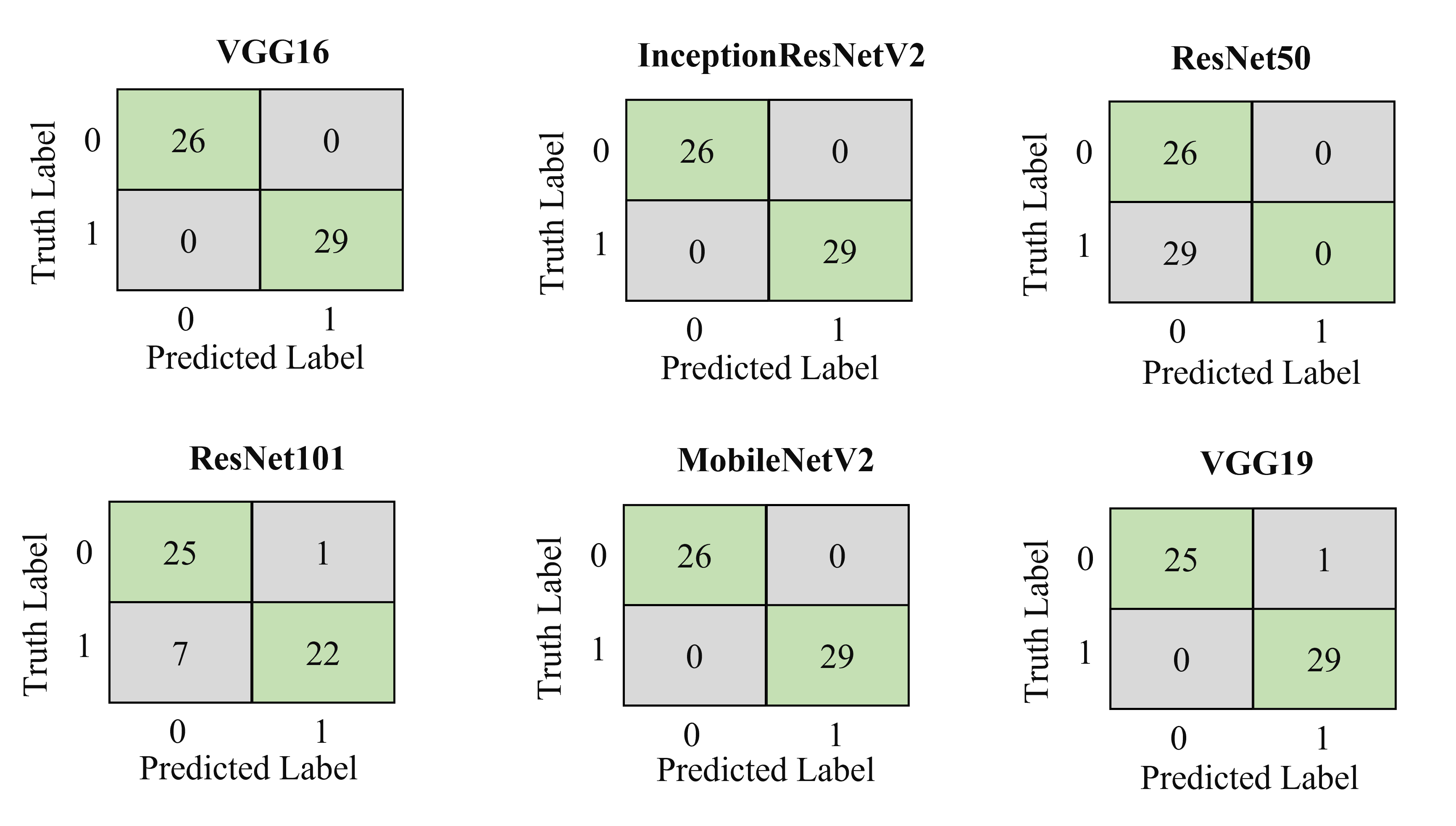}
    \caption{Confusion matrices of the different Transfer Learning models used during Study One on the test set.}
    \label{fig:cm22a}
\end{figure}

Figure~\ref{fig:cm22a} shows the confusion matrix on the test data for Study One. The algorithms VGG16, InceptionResNetV2, and MobileNetV2 achieved perfect accuracy with true positives (TP) of 26 and no false positives (FP) or false negatives (FN), while ResNet50 had a TP of 26 but a high number of false negatives (FN) at 29. ResNet101 had a slightly lower performance than VGG16, InceptionResNetV2, and MobileNetV2, with a TP of 25 and 1 FP, and 7 FN. VGG19 had a TP of 25 and 1 FP but no FN. 

\begin{figure}[H]
    \centering
    \includegraphics[width=\textwidth]{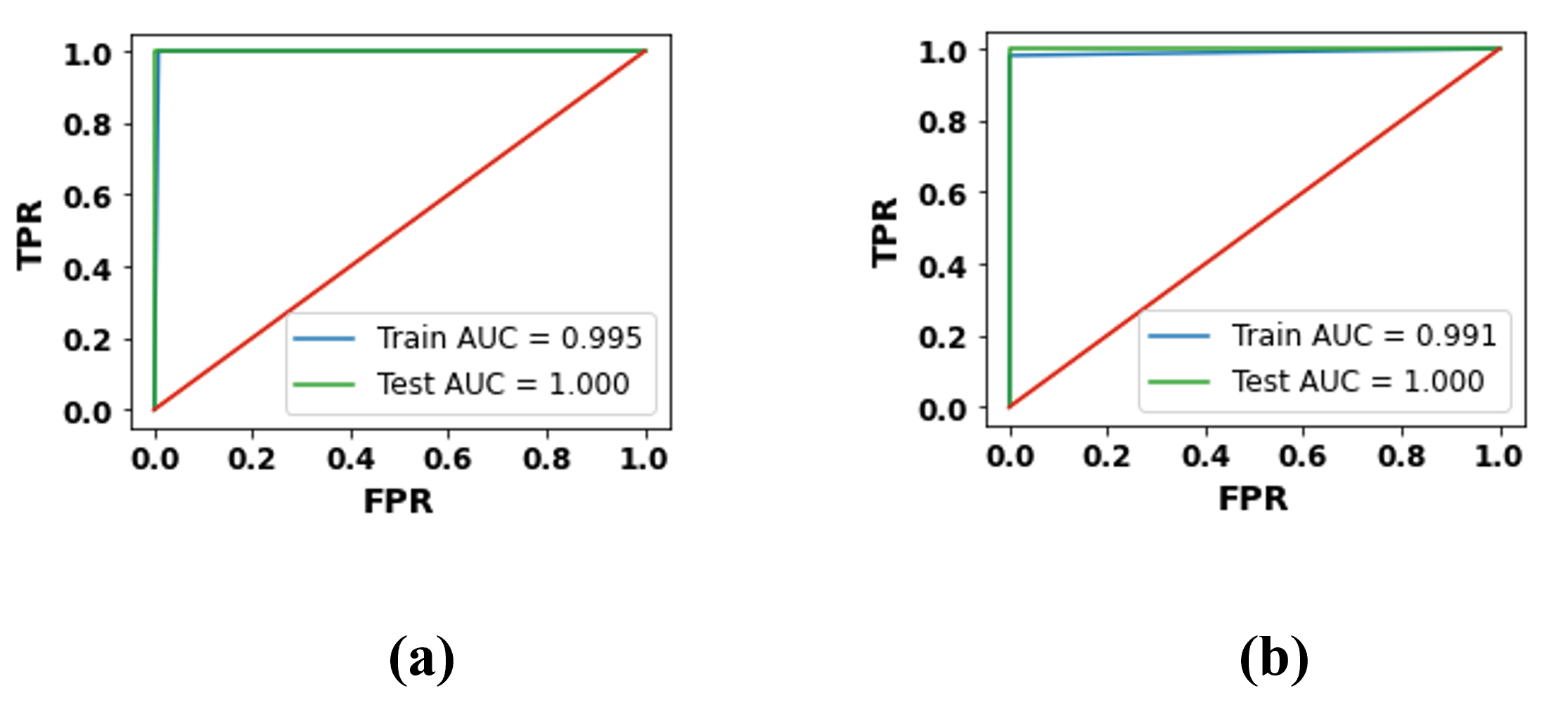}
    \caption{AUC score of two well-performed TL models (a) VGG16 and (b) InceptionResNetV2.}
    \label{fig:auc1}
\end{figure}

Figure~\ref{fig:auc1} demonstrates the top two DL models' AUC scores on the test set.

\begin{figure}[H]
    \centering
    \includegraphics[width=\textwidth]{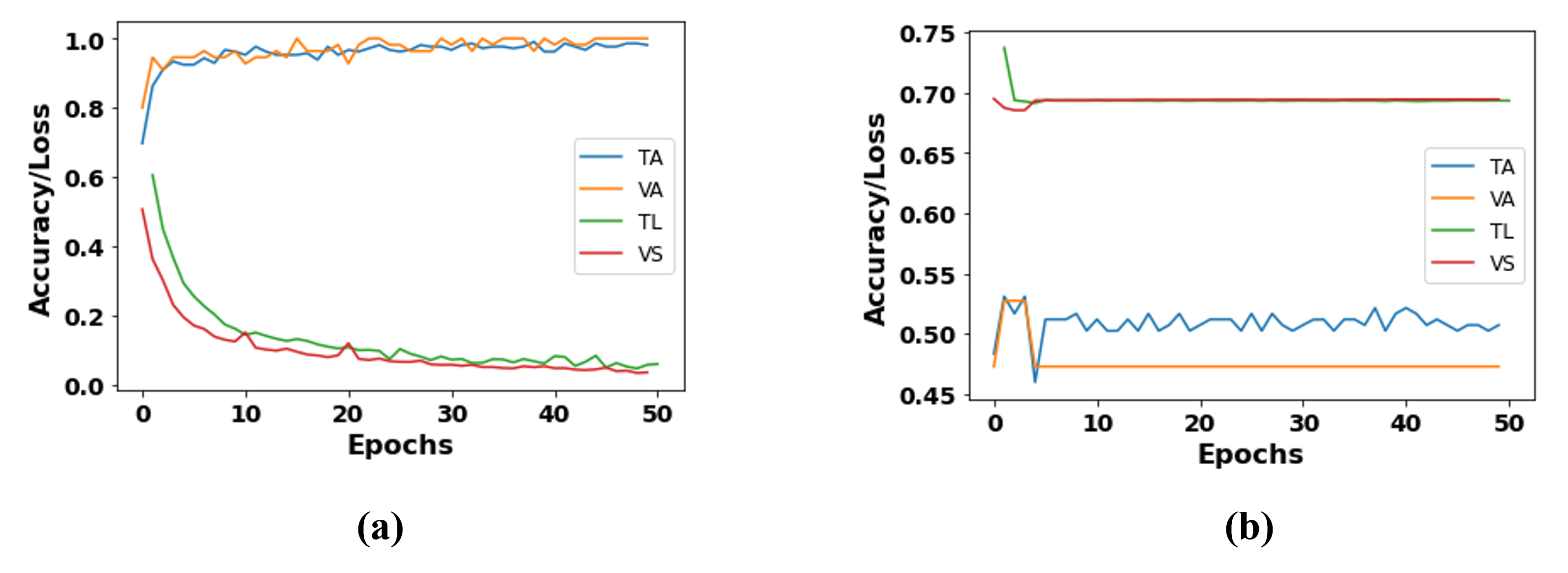}
    \caption{Training and validation performance throughout the training of (a) VGG16 and (b) ResNet50 during Study One. TA -- training accuracy; VA -- validation accuracy; TL -- training loss; VS -- validation loss.}
    \label{fig:TL2}
\end{figure}

Figure~\ref{fig:TL2} compares the In Figure~\ref{fig:TL2}, the performance of VGG16 and ResNet50 models during the training phase is compared. The results show that VGG16 continuously improved train accuracy and reduced train loss until epoch 20. At the same time, ResNet50 stopped showing any further improvements in performance after only ten epochs.

\subsection{Study Two}
Table~\ref{tab:std2train} presents the results of Study Two model performance on the train set. The best-performing models on the train set in Study Two are VGG16, InceptionResNetV2, MobileNetV2, and VGG19, which achieved perfect scores on all evaluation metrics. The worst-performing model is ResNet50, which achieved the lowest accuracy, precision, recall, and F1-score scores. ResNet101 showed overall good performance, with a score of 0.85 for accuracy and F1-score and precision and recall scores of 0.87 and 0.78, respectively.
\begin{table}[H]
\centering
\caption{Computational results of the various Transfer Learning models used in Study Two on the train set.}
\label{tab:std2train}
\resizebox{\textwidth}{!}{%
\begin{tabular}{@{}lcccc@{}}
\toprule
Algorithm & \multicolumn{1}{l}{Accuracy} & \multicolumn{1}{l}{Precision} & \multicolumn{1}{l}{Recall} & \multicolumn{1}{l}{F1-score} \\ \midrule
VGG16 & 1 & 1 & 1 & 1 \\
InceptionResNetV2 & 1 & 1 & 1 & 1 \\
ResNet50 & .8 & .89 & .69 & .71 \\
ResNet101 & 0.85 & 0.87 & 0.78 & 0.8 \\
MobileNetV2 & 1 & 1 & 1 & 1 \\
VGG19 & 1 & 1 & .99 & 1 \\ \bottomrule
\end{tabular}%
}
\end{table}

\begin{figure}
    \centering
    \includegraphics[width=\textwidth]{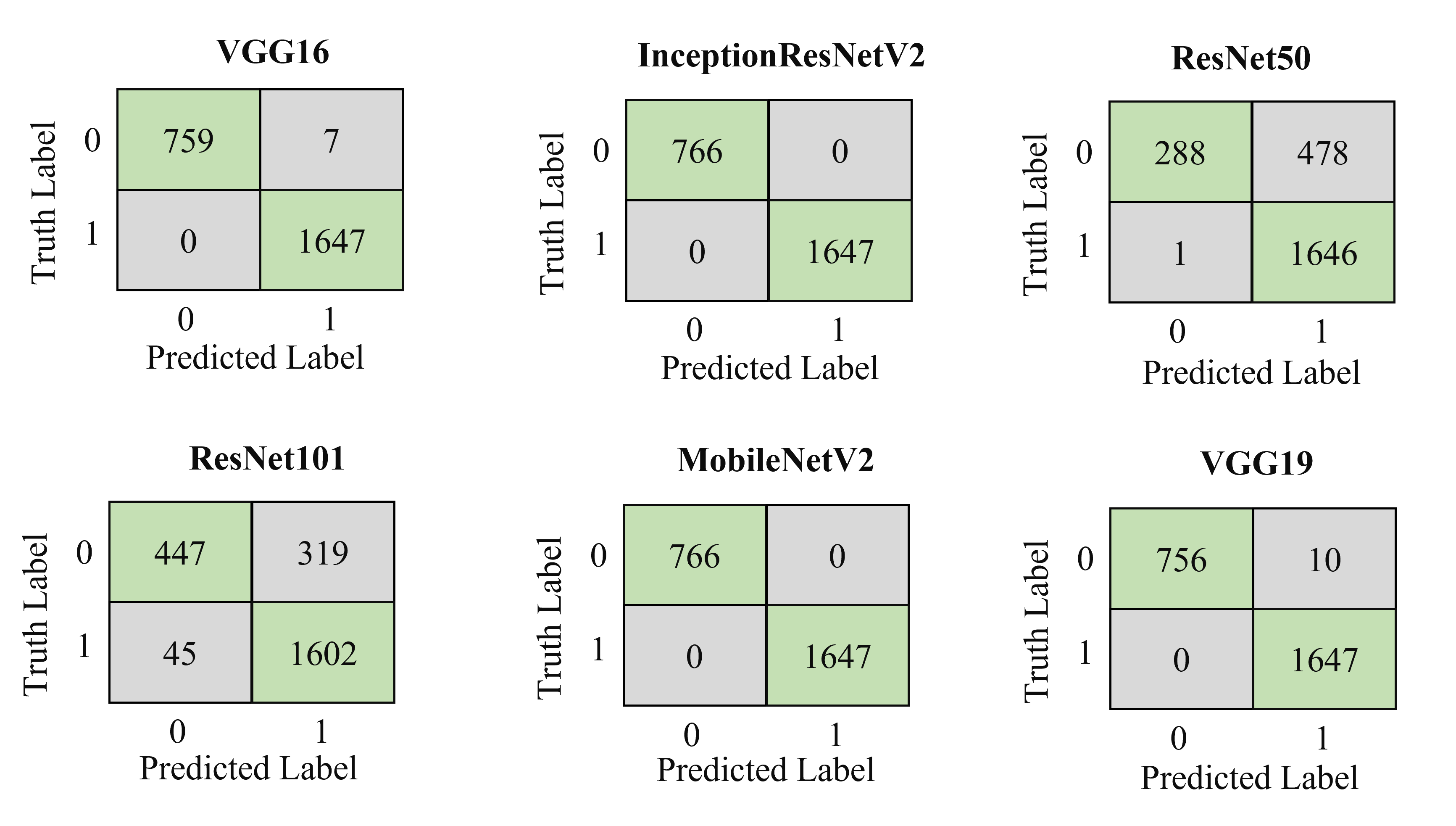}
    \caption{Confusion matrices of the different Transfer Learning models used during Study Two on the train set.}
    \label{fig:cm4a}
\end{figure}

Figure~\ref{fig:cm4a} shows the confusion matrix on the train data for Study Two. 
Among the algorithms, InceptionResNetV2 and MobileNetV2 had the highest number of TP, 766 each, while ResNet50 had the lowest, with only 288 TP. InceptionResNetV2 had the fewest FP, with 0, while ResNet50 had the most, with 478. ResNet101 had the highest number of FN, with 45, while VGG16, InceptionResNetV2, MobileNetV2, and VGG19 had none.

Overall, InceptionResNetV2 and MobileNetV2 achieved the best performance, with high numbers of TP and TN and no FN or FP. ResNet50 had the lowest performance, with a low number of TP and a high number of FP.

Table~\ref{tab:std2test} presents the results of Study Two model performance on the test set. The MobileNetV2, VGG16, InceptionResNetV2, and VGG19 models achieved the best results, with perfect accuracy, precision, recall, and F1-score. The ResNet50 and ResNet101 models showed lower performance in terms of accuracy, precision, recall, and F1-score, with the ResNet50 performing the worst among all models. 

\begin{table}[H]
\centering
\caption{Computational results of the various Transfer Learning models used in Study Two on the test set.}
\label{tab:std2test}
\resizebox{\textwidth}{!}{%
\begin{tabular}{@{}lcccc@{}}
\toprule
Algorithm & \multicolumn{1}{l}{Accuracy} & \multicolumn{1}{l}{Precision} & \multicolumn{1}{l}{Recall} & \multicolumn{1}{l}{F1-score} \\ \midrule
VGG16 & 1 & 1 & 1 & 1 \\
InceptionResNetV2 & 1 & 1 & 1 & 1 \\
ResNet50 & 0.83 & 0.9 & 0.72 & 0.75 \\
ResNet101 & 0.86 & 0.88 & 0.79 & 0.81 \\
MobileNetV2 & 1 & 1 & 1 & 1 \\
VGG19 & 1 & 1 & .99 & 1 \\ \bottomrule
\end{tabular}%
}
\end{table}

\begin{figure}[H]
    \centering
    \includegraphics[width=\textwidth]{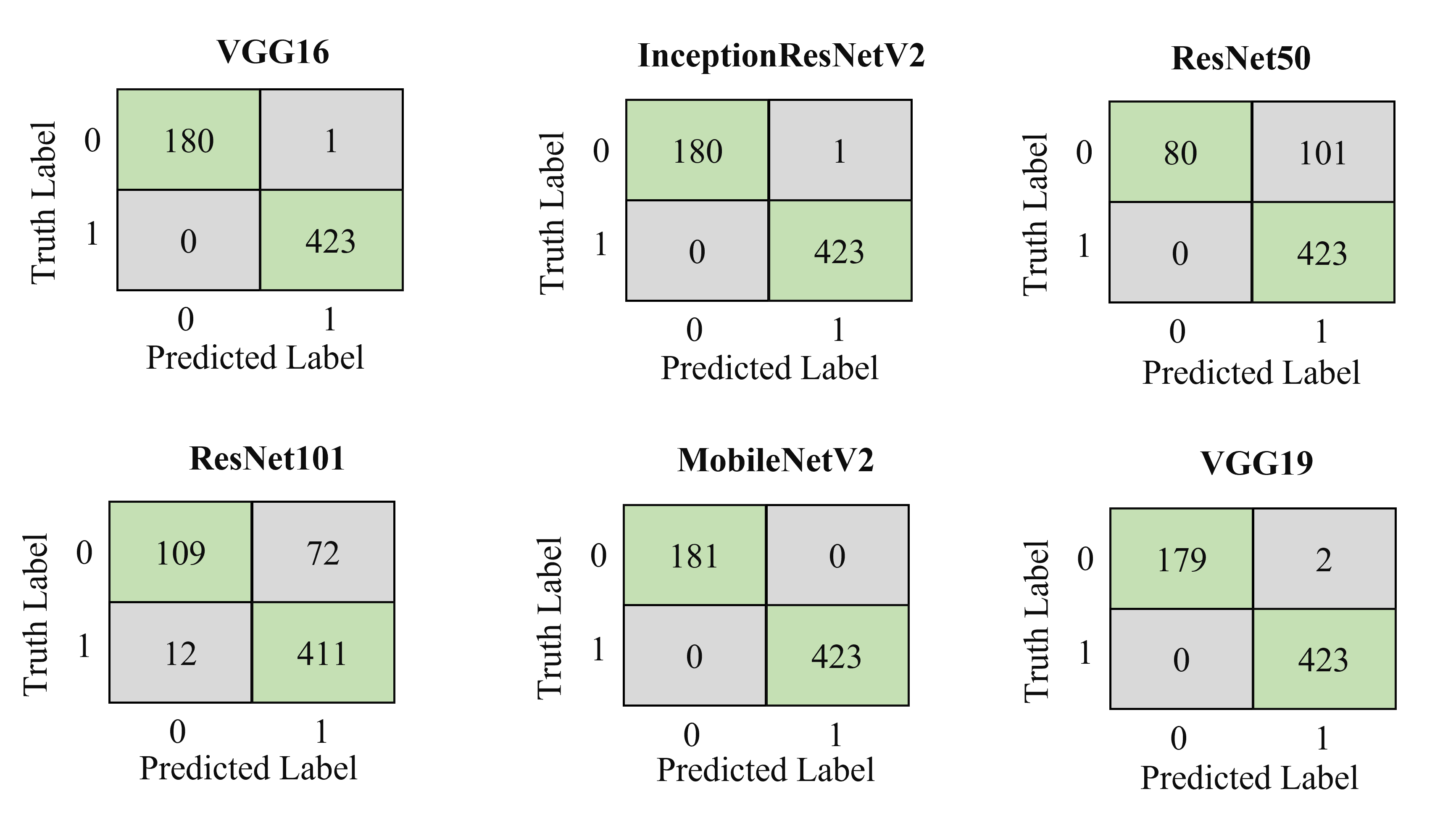}
    \caption{Confusion matrices of the different Transfer Learning models used during Study Two on the test set.}
    \label{fig:cm2a}
\end{figure}

Figure~\ref{fig:cm2a} shows the confusion matrix on the test data for Study Two. The figure shows that MobileNetV2 had the highest TP and TN, equaling 181, indicating that it correctly classified all positive and negative instances. VGG16 and InceptionResNetV2 also performed well, with TP and TN equaling 180 for each algorithm. ResNet50 had the highest number of FP, with 101, and the lowest number of TP, with only 80. ResNet101 had a lower number of FP (72) but a higher number of FN (12) than other models. VGG19 had the second-highest number of TN, equaling 423, but had 2 FP. Overall, MobileNetV2 showed the best performance, while ResNet50 and ResNet101 had relatively lower performance than other algorithms.

\begin{figure}[H]
    \centering
    \includegraphics[width=\textwidth]{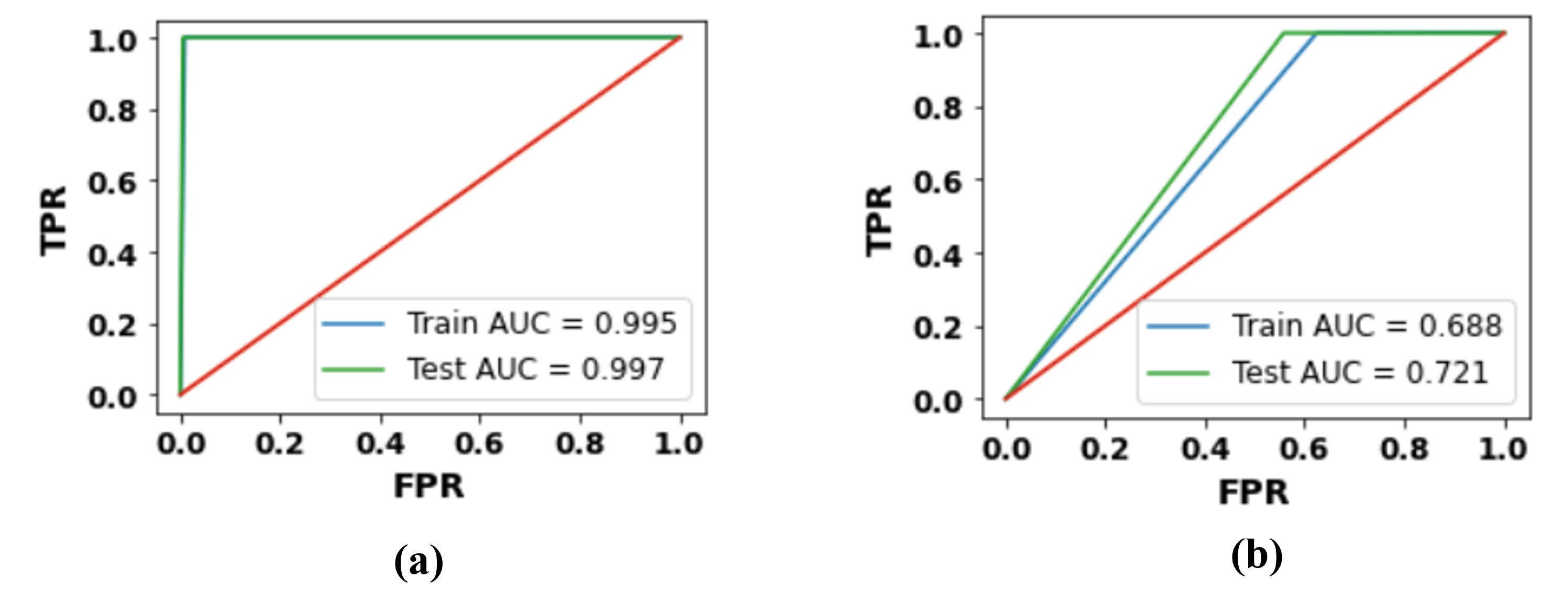}
    \caption{AUC score of (a) VGG16 and (b) ResNet50.}
    \label{fig:auc2}
\end{figure}

\begin{figure}[H]
    \centering
    \includegraphics[width=\textwidth]{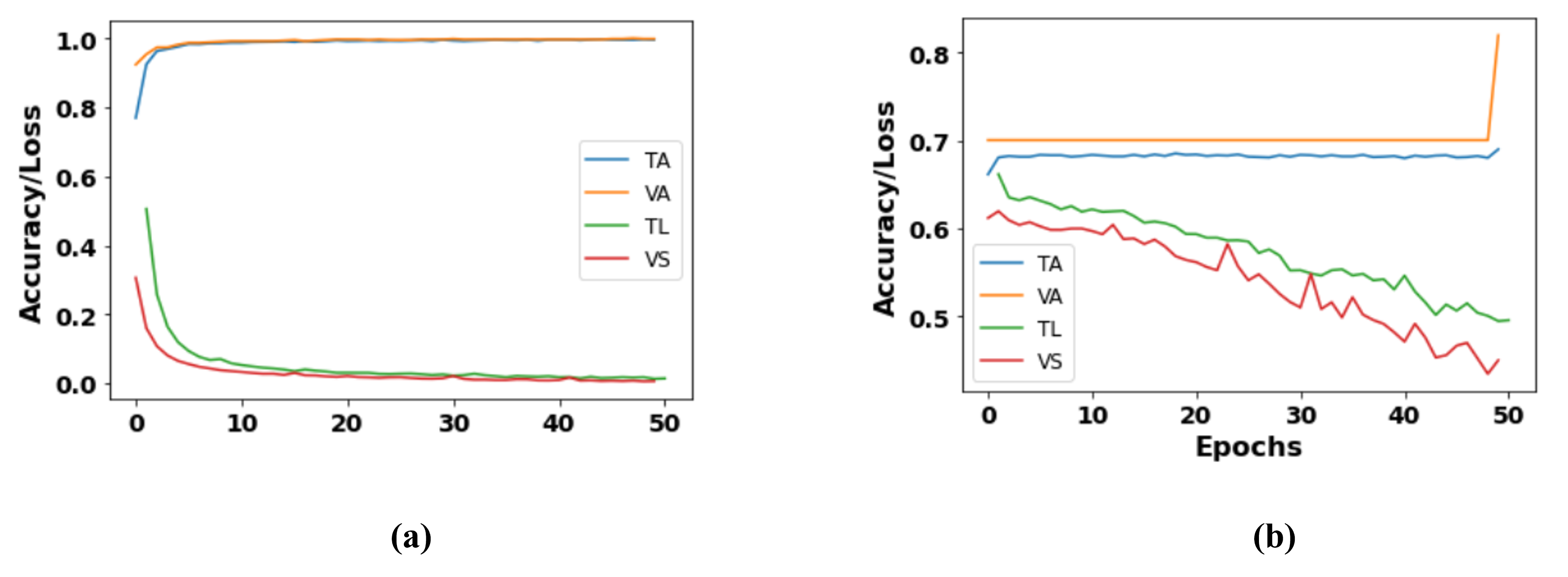}
    \caption{Training and validation performance throughout the training of (a) VGG16 and (b) ResNet50 during Study Two. TA -- training accuracy; VA -- validation accuracy; TL -- training loss; VS -- validation loss.}
    \label{fig:TL1}
\end{figure}

Figure~\ref{fig:auc2} illustrates the AUC score of VGG16 and ResNet50 models during Study Two. From the figure, it can be observed that the performance of VGG16 is nearly perfect, whereas the performance of ResNet15 demonstrates significantly poor performance.
Figure~\ref{fig:TL1} compares the performance of VGG16 and ResNet50 models during the training phase. The figure indicates that VGG16 constantly improved train accuracy and reduced train loss until epoch 20, whereas ResNet50 ceased showing further performance improvements after only ten epochs.

\section{Discussions}\label{discussion}
The current study investigated how deep learning approaches impact imbalanced data analysis. Six deep learning algorithms - VGG16, InceptionResNetV2, ResNet50, ResNet101, MobileNetV2, and VGG19 - were evaluated on an imbalanced data classification task. The results showed that VGG16 and MobileNetV2 achieved perfect scores on all four evaluation metrics, while InceptionResNetV2 performed well, averaging 99\% across all metrics. In contrast, ResNet50 had the lowest performance among the models, with an average F1-score of 0.34.

The confusion matrix analysis further supported these findings, showing that VGG16, InceptionResNetV2, and MobileNetV2 had perfect classification accuracy on the train data. In contrast, ResNet50 and VGG19 had lower accuracy due to higher false negative and false positive values. Moreover, on the test data for Study Two, MobileNetV2 demonstrated the best performance, correctly classifying all positive and negative instances. ResNet50 had the highest number of false positives and the lowest number of true positives, while ResNet101 had a higher number of false negatives.

Overall, the findings suggest that deep learning algorithms can perform well on imbalanced data classification tasks, but their performance may vary depending on the specific algorithm and the imbalance level. While some algorithms, such as MobileNetV2, can achieve high classification accuracy, others, such as ResNet50, may struggle with such tasks due to higher false positive and false negative rates. The limitations of this study include the use of a specific dataset and an imbalanced data classification task, which may not generalize to other datasets or tasks. Future research could investigate the impact of various DL algorithms on different imbalanced data classification tasks and explore ways to optimize their performance.
\subsection{Models Prediction}
\begin{figure}[H]
    \centering
    \includegraphics[width=\textwidth]{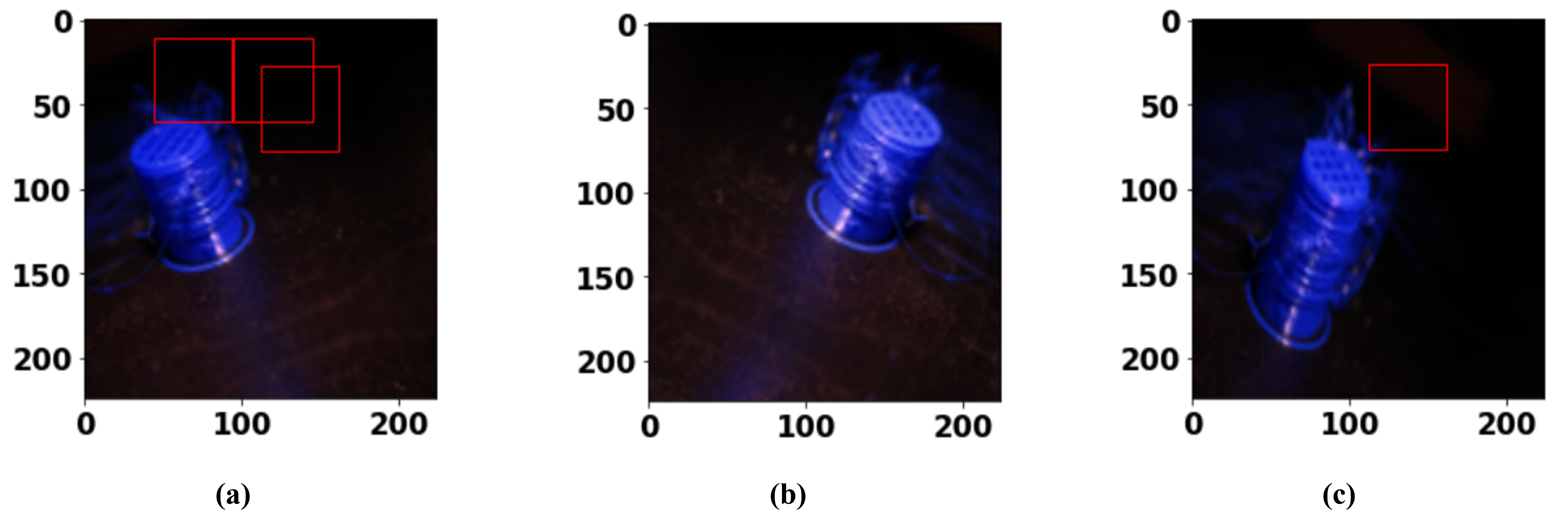}
    \caption{Failure to accurately locate the defect in the 3D printed cylinder of (a) VGG16, (b) ResNet50, and (d) MobileNetV2.}
    \label{fig:fail}
\end{figure}

Figure~\ref{fig:fail} illustrates the model's failure to detect the defect in the image of the 3D-printed cylinder. Based on the results presented in Figure~\ref{fig:fail}, it is evident that all three models failed to locate the defect in the 3D-printed cylinder accurately. Although VGG16 and MobileNetV2 produced bounding boxes that were relatively close to the defect, they were still unable to localize it accurately. These models show little promise and could be used for further investigation. However, similar poor performance was also observed for ResNet101, InceptionResNetV2, and VGG19, indicating that these models may not be well-suited for defect localization in 3D printed objects. These differences in performance may be attributed to variations in the models' architecture and how they handle spatial features. Nonetheless, it is essential to note that despite these differences, all models failed to locate the defect accurately, highlighting the challenges associated with defect localization in 3D printed objects. These findings emphasize the need for further investigation into the factors contributing to model failures and suggest that optimizing model architecture and training strategies may improve localization accuracy and develop more effective solutions for defect detection and classification in additive manufacturing processes.


\section{Conclusion}\label{conclusions}
The present study investigated the impact of DL-based approaches on imbalanced data analysis using six popular TL-based DL methods, including VGG16, InceptionResNetV2, ResNet50, MobileNetV2, ResNet101, and VGG19. The experiments were conducted on both small balanced and large imbalanced datasets containing images of the 3D-printed cylinder. 

Based on the experiment, it can be answered that even though the performance of TL-based methods on balanced and imbalanced data is very high, the existing popular TL-based approaches failed to locate the potential area, such as defect regions in cylinder images, which indicates that imbalanced data may affect TL-based models' performance. Additionally, VGG16 and MobileNetV2 showed
promising results and are a good choice for further investigation.

However, additional experiments with larger and different
datasets are necessary to validate the findings. In future work, it is planned to conduct experiments with various datasets and propose advanced TL-based approaches that may help overcome the current challenges observed during this experiment. Ultimately, the study highlights the need for further investigation into the factors contributing to model failures and developing more effective solutions for defect detection and classification in additive manufacturing processes.

\bibliographystyle{unsrt}  
\bibliography{main}

\end{document}